\pdfoutput=1

\documentclass[11pt]{article}

\usepackage[preprint]{coling}

\usepackage{times}
\usepackage{latexsym}

\usepackage[T1]{fontenc}

\usepackage[utf8]{inputenc}

\usepackage{microtype}

\usepackage{inconsolata}

\usepackage{graphicx}
\usepackage{booktabs}
\usepackage{multirow}

\usepackage{natbib}

\title{Task-Oriented Dialog Systems for the Senegalese Wolof Language}

\author{Derguene Mbaye\textsuperscript{1,2} \\
  \textsuperscript{1}Baamtu Technologies \\
  \textsuperscript{2}Universit\'{e} Cheikh Anta Diop \\ Dakar, S\'{e}n\'{e}gal \\
  \texttt{derguenembaye@esp.sn} \\\And
  Moussa Diallo\textsuperscript{3} \\
  \textsuperscript{3}Ecole Sup\'erieure Polytechnique (ESP) \\ Dakar, S\'{e}n\'{e}gal \\
  \texttt{moussa.diallo@esp.sn} \\}

\begin{document}
\maketitle
\begin{abstract}
In recent years, we are seeing considerable interest in conversational agents with the rise of large language models (LLMs). Although they offer considerable advantages, LLMs also present significant risks, such as hallucination, which hinder their widespread deployment in industry. Moreover, low-resource languages such as African ones are still underrepresented in these systems limiting their performance in these languages. In this paper, we illustrate a more classical approach based on modular architectures of Task-oriented Dialog Systems (ToDS) offering better control over outputs. We propose a chatbot generation engine based on the Rasa framework and a robust methodology for projecting annotations onto the Wolof language using an in-house machine translation system. After evaluating a generated chatbot trained on the Amazon Massive dataset, our Wolof Intent Classifier performs similarly to the one obtained for French, which is a resource-rich language. We also show that this approach is extensible to other low-resource languages, thanks to the intent classifier's language-agnostic pipeline, simplifying the design of chatbots in these languages.
\end{abstract}

\section{Introduction}\label{sec:introduction}
Artificial Intelligence (AI) has experienced a tremendous growth in recent years, mainly due to the rapid development of Deep Learning.  The latter has allowed to achieve super-human skills on tasks such as Image Classification. This has been mainly made possible thanks to the development of large datasets and the considerable increase in computing power and their accessibility. NLP has thus leveraged these advances and has had its "ImageNet moment"\footnote{\underline{https://www.ruder.io/nlp-imagenet/}} with the emergence of huge corpora as well as the possibility of leveraging pre-trained models to apply them to downstream tasks through Transfer Learning \cite{ruder-etal-2019-transfer}.

However, the availability of such corpora only concerns a small group of languages such as English, Chinese or French which are referred to as "Resource Rich". The majority of the approximately remaining 7,000 languages \cite{ethnologue-2019}, and particularly African ones, fall into the "Low Resource" category and struggle to have sufficient corpora and NLP tools \cite{hedderich2021survey}. They are thus left behind in most of the AI revolutions, such as LLMs, which reach the state of the art in many NLP tasks but struggle to reproduce equivalent performance in African languages. A human-translated benchmark dataset for 16 typologically diverse low-resource African languages (including Wolof) has been presented by \cite{adelani2024irokobenchnewbenchmarkafrican} covering three tasks: natural language
inference, mathematical reasoning, and multi-choice knowledge-based QA. They evaluated zero-shot, few-shot, and translate-test settings (where test sets are translated into English)
across 10 open and four proprietary LLMs and revealed a significant performance gap between high-resource languages (English
and French) and African ones. A lot of work has therefore been done to address this challenge and a great illustration is the emergence of initiatives such as the  Masakhane\footnote{\underline{https://www.masakhane.io/}} community which brings together thousands of researchers, practitioners, linguists and enthusiasts to produce datasets and models for a wide range of African languages \cite{nekoto2020participatory}.
Other data collection projects have also included African languages, and especially Wolof, such as \cite{TIEDEMANN12.463, strassel-tracey-2016-lorelei, goyal-etal-2022-flores, federmann-etal-2022-ntrex}. 

However, the data collection process is a very time-consuming task and can quickly become tedious. Additionally, for some languages, the text may not even exist in electronic form, or worse still, the text may be obtained from phonetic transcriptions directly from audio, in the absence of a defined written standard. So, it is important to point out here that not all low-resource languages are at the same level in terms of resource scarcity. Other initiatives have therefore studied the usability of work already done in resource-rich languages in order to apply it to low resource ones. This is particularly the case with Cross-Lingual Transfer which is an approach where knowledge gained while training a model on a particular language (or set of languages) is leveraged to improve the performance on tasks in a different one  \cite{artetxe-etal-2020-cross, lample2019crosslingual}. This state-of-the-art technique is applicable to virtually all natural language processing (NLP) tasks, including Task-oriented Dialog Systems (ToDS) where the lack of datasets in most languages for both training and evaluation is the most critical factor preventing the creation of truly multilingual ToDS \cite{razumovskaia2022crossingconversationalchasmprimer}. ToDS are software programs (or agents) that users talk to in order to carry out a given task. To do this, they usually start by identifying the user's goal, often called intent, and the associated arguments, called slots, before being able to process the request appropriately. The most common applications are hotel, taxi- and restaurant reservations, information supply and customer support, among others. In the following example: \textbf{Book me a room for a person from July 15 to July 24}, the intent would be \underline{book\_room} and the \textit{start\_date} and \textit{end\_date} slots would have the values \textit{July 15} and \textit{July 24} respectively. Intent recognition and slot filling are thus the main tasks involved in a ToDS, and can be done either separately \cite{arora2020crosslingual} or jointly \cite{schuster-etal-2019-cross-lingual} in a single task. This architecture, in which the system is divided into several modules performing specific tasks, is called modular architecture as opposed to the end-to-end one which uses a neural network for all these tasks \cite{zhang2020recentadvanceschallengestods}. This makes them very data-intensive, as they need to model several different tasks from training data.

In this paper, we propose to leverage the benefits of Cross Lingual Transfer to build a Wolof Chatbot generation engine based on an in-house French$\leftrightarrow$Wolof machine translation system and a language agnostic pipeline for intent classification and slot filling using Rasa \cite{bocklisch2017rasaopensourcelanguage}. We introduce a simple but effective approach to projecting annotations from a source language to a target one, using only the original machine translation system coupled with an intuitive parsing strategy. The paper is therefore structured as follows:
\begin{itemize}
  \item We begin by presenting some work done in Cross-lingual Transfer for Task oriented Dialog Systems on low-resource languages in Section \ref{rel-work}.
  \item The Wolof language and the datasets used are presented in Section \ref{data}.
  \item Our annotation projection approach is presented in section \ref{projection}.
  \item In Section \ref{exp}, we present the different experiment settings.
  \item Section \ref{results} presents the results.
  \item Conclusion and perspectives are presented in section \ref{concl}.
\end{itemize}

\section{Related Work}\label{rel-work}
ToDS for low-resource languages has been the subject of extensive research in the community and various approaches have been studied. A first experience of a Wolof chatbot was proposed in \cite{gauthier-etal-2022-preuve} with a POC of a voice assistant based on Rasa and designed with manually collected synthetic data. This limits the ability to collect large volumes of data, which is the reason why more efficient alternatives have been explored.

\subsection{Word alignment approach}
Training data translation (translate-train) into target languages is an approach that is increasingly being considered to enhance the performance of cross-lingual transfer. \cite{schuster-etal-2019-cross-lingual} explored three different cross-lingual transfer methods: translation of training data with a bidirectional neural machine translation system (NMT) following the work carried out by \cite{mccann2018learnedtranslationcontextualizedword}, using cross-lingual pre-trained embeddings, and a method of using a multilingual machine translation encoder as contextual word representations. Although the latter two showed better performance in the presence of hundreds of training data, in cases where no data is available in the target language, translating the training data gives the best results. This translation approach has also been explored by \cite{lopez-de-lacalle-etal-2020-building}, which leveraged the Spanish subset of the dataset of \cite{schuster-etal-2019-cross-lingual} to translate it into Basque using a Spanish-to-Basque Transformer-based NMT system \cite{vaswani2023attentionneed} before projecting the annotations using word alignment \cite{dyer-etal-2013-simple}. Two additional models were used to perform the word alignment, making this approach more laborious. A similar approach was also presented by \cite{kanakagiri-radhakrishnan-2021-task} with an NMT-based dataset creation approach and an annotation projection based on token prefix matching and mBERT \cite{devlin-etal-2019-bert} based semantic matching from a rich resource language. For the translation phase, they also experimented with a transformer-based model but also with Google Translate services.

\subsection{Marker-based approach}
To avoid the need for additional models for word alignment, other simpler alternatives are considered, such as mark-then-translate as explored in \cite{lewis-etal-2020-mlqa}. This approach consists of inserting specific markers around the corresponding chunks in the source sentence (e.g., [marker] and [/marker]), then translating into the target language to maintain these markers during the translation process. An optimized version of the mark-then-translate approach, has been proposed in \cite{chen-etal-2023-frustratingly} consistently outperforming the alignment-based approach. Authors used language-agnostic square bracket markers, combined with an efﬁcient ﬁne-tuning strategy of the NLLB (No Language Left Behind) model \cite{nllbteam2022language} to encourage the multilingual machine translation system to better preserve the special markers during translation. They also showed that the alignment-based methods are more error-prone when projecting span-level annotations, compared to the marker-based approaches. However, inserting markers into the source sentence tends to compromise the translation quality \cite{chen-etal-2023-frustratingly}. A different approach called Translate-and-Fill (TaF) is explored in \cite{nicosia-etal-2021-translate-fill}, as opposed to Translate-Align-Project (TAP), which uses alignment and projection modules. TaF consists of a sequence-to-sequence ﬁller model that constructs a full parse conditioned on an utterance and a view of the same parse. This approach requires however two seq2seq models trained differently: one is the usual semantic parser and the other is what they call the ﬁller. Researchers in \cite{xu-etal-2020-end} introduced a single end-to-end model that learns to align and predict target slot labels jointly for cross-lingual transfer.

\section{Data}\label{data}

\subsection{The Wolof Language}
Belonging to the Atlantic group of
the Niger-Congo language family, Wolof is an African language mainly spoken in Senegal (the majority of the population) but also in Gambia and some parts of Mauritania. It is a non-tonal agglutinative language whose alphabet is quite close to the French one: we can ﬁnd all the letters of its alphabet except H, V and Z \cite{derguenelstm}. As with many African languages, Wolof is a low-resource language, as opposed to high-resource languages like English and Chinese. Although there is no consensus on a suitable definition of the low-resource concept, several researchers have explored definitions from different angles even beyond the lack of data and NLP tools. Thus, this concept has been defined in \cite{asrlrl} as a language that lacks a unique writing system, has a low Internet presence, and lacks linguistic expertise, among other things. Wolof linguistics has, however, benefited from a lot of research but the lack of a unique writing system makes the design of NLP tools particularly challenging in this language. There is a writing system based on the Latin script and another one based on the Arabic script called Wolofal or Ajami \cite{derguenelstm}. The latter is however very little used and is generally located within usage related to the Islamic religion which is the predominant religion in Senegal. The Latin script, on the other hand, is the most widespread, and is the one we'll be considering in this work. However, due to a lack of language standardization, two forms of writing illustrated in \cite{mbaye2023beqirevitalizesenegalesewolof} are noted with the Latin script:
\begin{enumerate}
  \item A so-called \texttt{official} form, respecting the published alphabet and adopted by all Wolof datasets open to date;
  \item A \texttt{conventional} form that doesn't respect any rules and is very present on media such as social networks.
\end{enumerate}
We'll therefore consider the \texttt{official} form in this article.

\subsection{Dataset}
There are several multilingual datasets in Task-oriented dialog systems, typically for intent classification and slot filling. Some datasets specialize in a particular domain, such as MultiATIS++ \cite{xu-etal-2020-end}, which is specific to air travel, while others are multi-domain, such as MTOP \cite{li-etal-2021-mtop}, which covers 11 domains. Among ToDS datasets, the MASSIVE dataset \cite{fitzgerald-etal-2023-massive} offers the greatest diversity in terms of languages and domains, with 51 languages and 18 domains covered, as illustrated in the Slot and Intent Detection benchmark and datasets presented in \cite{kwon-etal-2023-sidlr}.
It also contains examples in French, which will enable us to use a French$\rightarrow$Wolof machine translation system to project annotations in Wolof. Containing 1M labelled realistic utterances and around 16,500 utterances in French, we extracted 10 domains out of 18 spread over 27 intents and around 10,000 utterances (examples), as illustrated in Table \ref{tab:dataset}. French is the easiest language to work with when it comes to leveraging a resource-rich language to perform NLP tasks on Wolof. Since the official language in Senegal is French, it will be easier to find translators and other linguists on the French-Wolof language pair, and most existing translation datasets are in this language pair too. Wolof spoken in Senegal is also very much code-switched with French, with many French words used in everyday conversations.

\begin{table}[ht]
\centering
\small
\setlength{\tabcolsep}{4pt}
\begin{tabular}{@{}llr@{}}
\toprule
\textbf{Domains} & \textbf{Intents} & \textbf{No examples} \\
\midrule
\multirow{4}{*}{transport} & transport\_query & 314 \\
 & transport\_ticket & 187 \\
 & transport\_taxi & 150 \\
 & transport\_traffic & 154 \\
\midrule
\multirow{3}{*}{calendar} & calendar\_query & 794 \\
 & calendar\_set & 1150 \\
 & calendar\_remove & 426 \\
\midrule
\multirow{3}{*}{alarm} & alarm\_set & 254 \\
 & alarm\_remove & 113 \\
 & alarm\_query & 183 \\
\midrule
\multirow{3}{*}{lists} & lists\_query & 299 \\
 & lists\_remove & 253 \\
 & lists\_createoradd & 241 \\
\midrule
\multirow{2}{*}{takeaway} & takeaway\_query & 181 \\
 & takeaway\_order & 177 \\
\midrule
\multirow{5}{*}{play} & play\_audiobook & 226 \\
 & play\_game & 169 \\
 & play\_music & 938 \\
 & play\_podcasts & 290 \\
 & play\_radio & 401 \\
\midrule
news & news\_query & 709 \\
\midrule
\multirow{3}{*}{recommendation} & recommendation\_locations & 235 \\
 & recommendation\_events & 259 \\
 & recommendation\_movies & 102 \\
\midrule
\multirow{2}{*}{datetime} & datetime\_query & 502 \\
 & datetime\_convert & 76 \\
\midrule
weather & weather\_query & 855 \\
\midrule
 & \textbf{Total No examples} & \textbf{9638} \\
\bottomrule
\end{tabular}
\caption{Description of the Massive dataset extract selected}
\label{tab:dataset}
\end{table}

\section{Annotation Projection}\label{projection}
Annotation projection involves transferring the labels present in a sentence in a source language to the translated sentence in the target language. A machine translation system is thus at the heart of the process, and the resource-limited nature of the Wolof language adds a further layer of complexity. A French Wolof machine translation system based on LSTMs was presented in \cite{derguenelstm}, and another one based on Vanilla Transformers in \cite{dione-etal-2022-low}. But pre-trained multilingual translation models have shown the most interesting performances in supporting low-resource languages.  They allow information sharing between similar languages that greatly allows to improve translation on the language pairs as studied in \cite{arivazhagan2019massively}. A lot of work has been done in this direction and a wide range of multilingual translation models have therefore been developed. Researchers in \cite{adelani-etal-2022-thousand} compared a set of multilingual models for machine translation in over twenty African languages, including Wolof, and the M2M100 \cite{fan2020englishcentric} model showed the best performance. A distilled version of this model has been proposed in \cite{mohammadshahi2022small100}, offering equivalent performance while being 3.6x smaller and 4.3x faster at inference. We finetuned this model simultaneously in both directions on an in-house dataset of 175,000 French-Wolof sentences  and obtained a BLEU score of \textbf{26.38}.

We then empirically tested a set of markers including \texttt{xml tags}, \texttt{dollars}, \texttt{braces}, \texttt{brackets}, \texttt{parentheses} and some \texttt{random special characters} on sample translations, but found a range of issues:
\begin{enumerate}
    \item These markers tend to change the meaning of the sentence ;
    \item Their consistency is poor; some markers are preserved during translation in some sentences, but not in others, with no apparent pattern ;
    \item When they are preserved, the labels surrounded by the markers are sometimes also translated, which we aim to avoid.
\end{enumerate}

\begin{figure}
\centerline{\includegraphics[width=0.5\textwidth, keepaspectratio]{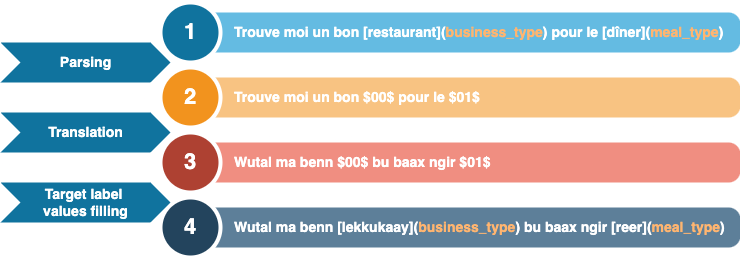}}
\caption{The three-step annotation projection algorithm: Parsing, which replaces the source annotations with ids, translation of the parsed sentence and backfilling of the translated annotations.}
\label{annotation}
\end{figure}

To address these challenges, we opted for \texttt{“identifiers”} instead of markers, by completely replacing the elements to be projected with numbered labels. Some numbers used as identifiers ended up being converted to letters during translation, reproducing the consistency issue mentioned above. We therefore combined the two approaches (markers and identifiers) by testing the markers presented earlier on the numbered identifiers. In the end, we found that only the dollar sign (\texttt{\$}) managed to preserve the marked identifiers during translation on all the sample sentences we tested. This may be due to the fact that this combination is the only one that is rare enough in the training data for the model not to confuse it and preserve it instead of trying to translate it in a certain way.
We thus replaced all the labeled words with an identifier of the form \texttt{\$0N\$}, where \texttt{N} is a number ranging from 0 to the total number of labeled terms in the dataset. The label of each replaced word is stored in a dictionary, as is the translation of the word in question. The resulting sentences were then passed to the translation system to obtain the equivalents in the target language (Wolof) while keeping the identifiers intact during the translation process. From there, the identifiers in the translated sentences are replaced back by the translation of the words stored in the dictionary with their labels. Thus, we obtain as output the translated sentences with their corresponding annotations in the target language. The overall procedure is illustrated in Fig.\ref{annotation}.

\section{Experiments}\label{exp}
\begin{figure}
\centerline{\includegraphics[width=0.5\textwidth, keepaspectratio]{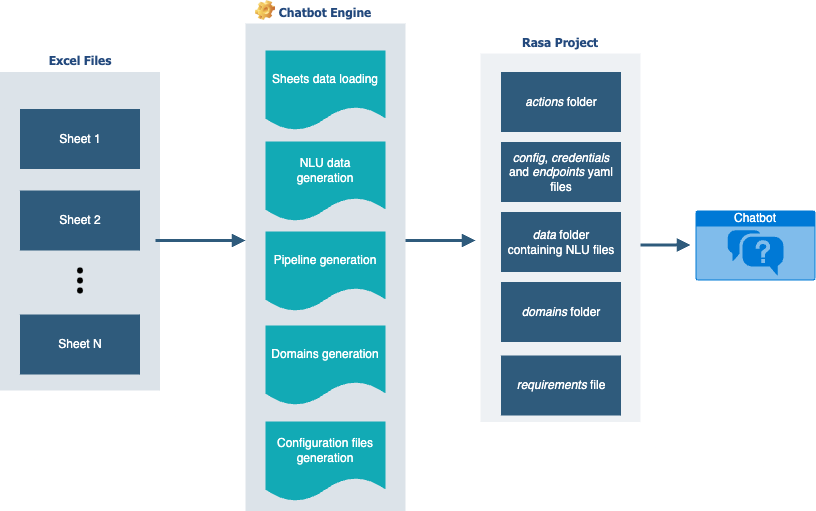}}
\caption{Diagram of the chatbot engine's processing of excel files to output Rasa projects. Each Excel file constitutes a domain containing several sheets corresponding to intents, and each sheet contains the intent's example data.}
\label{engine}
\end{figure}

Instead of directly building an intent classifier and slot filling, we propose a system for generating chatbots on the fly (chatbot engine), based on the Rasa framework \cite{bocklisch2017rasaopensourcelanguage}. Rasa is an open-source framework designed for creating conversational AI chatbots. It offers tools and libraries for building and deploying AI-powered, text and voice based chatbots capable of engaging in natural language conversations with users. The chatbot engine takes as input one or more Excel files which form the bot's ontology (domain, nlu and dialog data), and then generate a working RASA Chatbot from this as illustrated in Fig.\ref{engine}.

\begin{figure}
\centerline{\includegraphics[width=0.45\textwidth, keepaspectratio]{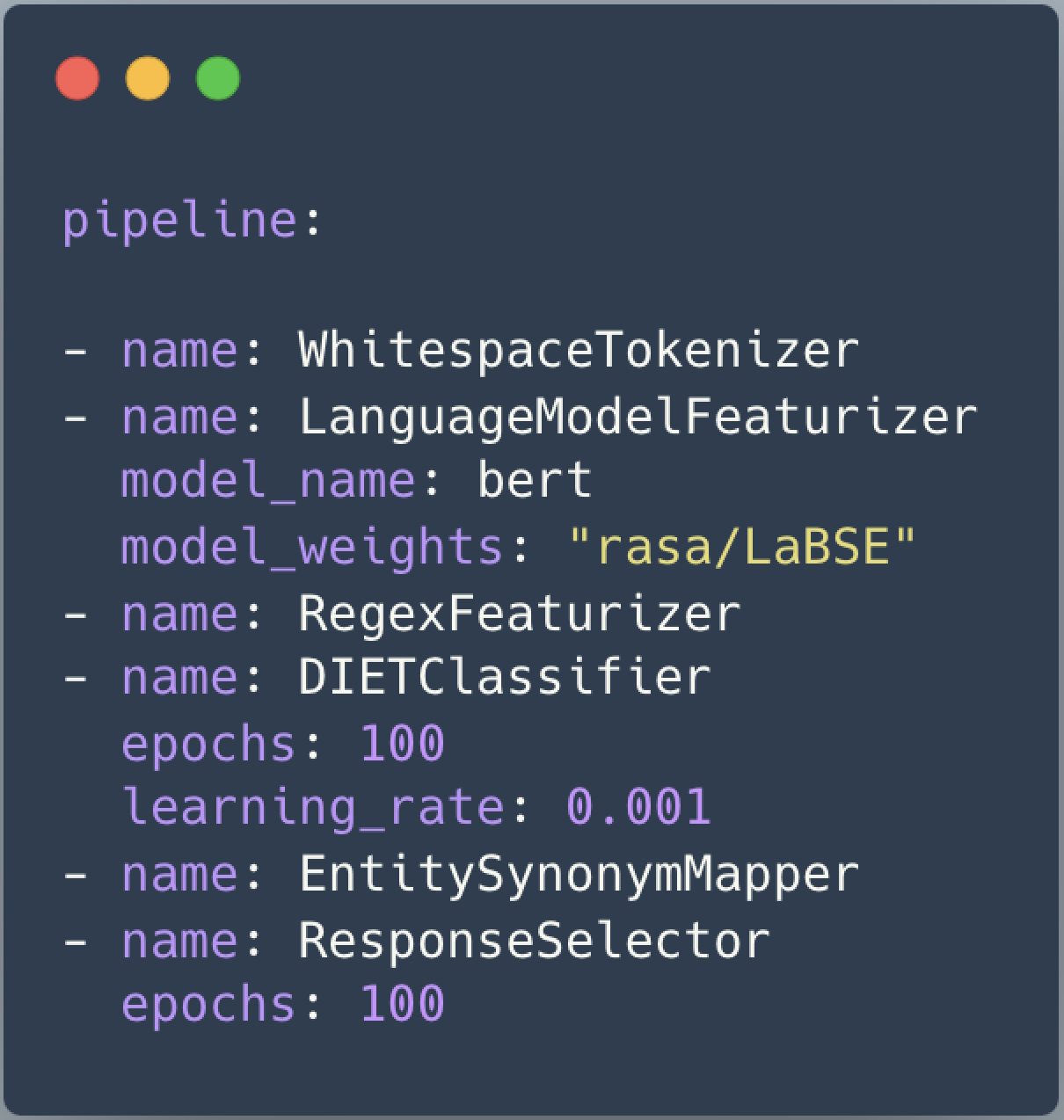}}
\caption{Pipeline of user input processing modules defined in the \texttt{config.yml} file generated by the chatbot engine.}
\label{pipeline}
\end{figure}

A benchmark for evaluating language-agnostic intent classiﬁcation has been studied in \cite{wang-etal-2022-benchmarking} and Language-agnostic BERT Sentence Embedding (LaBSE) \cite{feng2022languageagnosticbertsentenceembedding} produced the highest accuracy in almost all evaluation settings. Despite the long training times involved, this model allows the chatbots designed on top of it to be particularly flexible in terms of language support. We've used it to propose a simple but highly effective fixed pipeline, as illustrated in Fig.\ref{pipeline}. 

This pipeline is generated by the chatbot engine regardless of the language used in the excel files, making the overall system scalable to other languages beyond Wolof. At the heart of the pipeline is DIET\footnote{Dual Intent and Entity Transformer} \cite{bunk2020dietlightweightlanguageunderstanding}, a state-of-the-art Rasa model for intent detection and slot filling. Since French is a resource-rich language, it is better represented in LaBSE than Wolof, enabling it to create richer embeddings and thus a more efficient intent classifier. We therefore use the French part of the dataset as a baseline to evaluate performance on synthetic data generated through translation and annotation projection. We then randomly split our dataset into train/test with a ratio of 80/20. The results are reported as F1 scores.

\section{Results} \label{results}
\begin{figure}
\centerline{\includegraphics[width=0.5\textwidth, keepaspectratio]{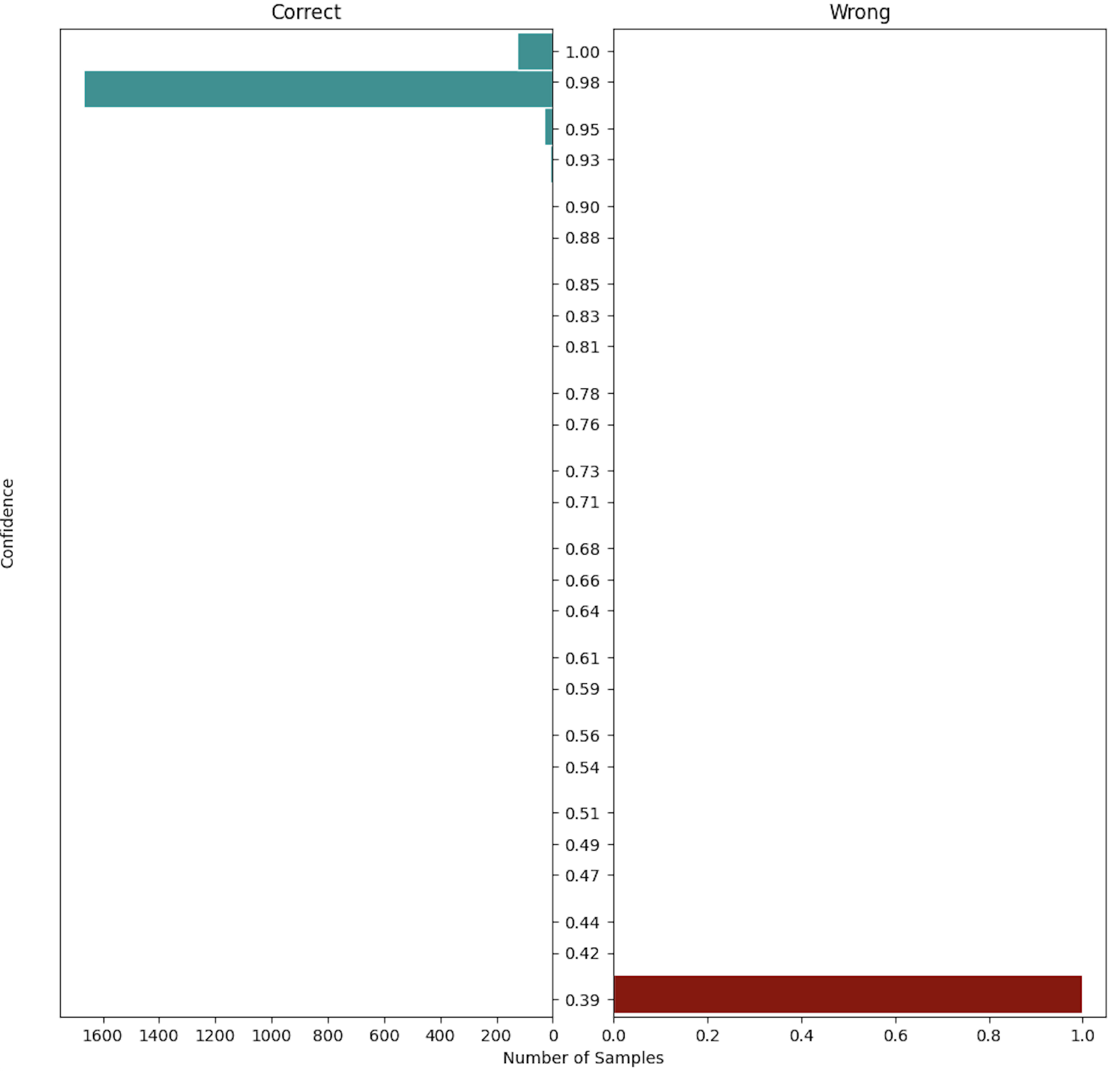}}
\caption{Intent prediction confidence distribution on the French dataset}
\label{french_intent}
\end{figure}

\begin{figure}
\centerline{\includegraphics[width=0.5\textwidth, keepaspectratio]{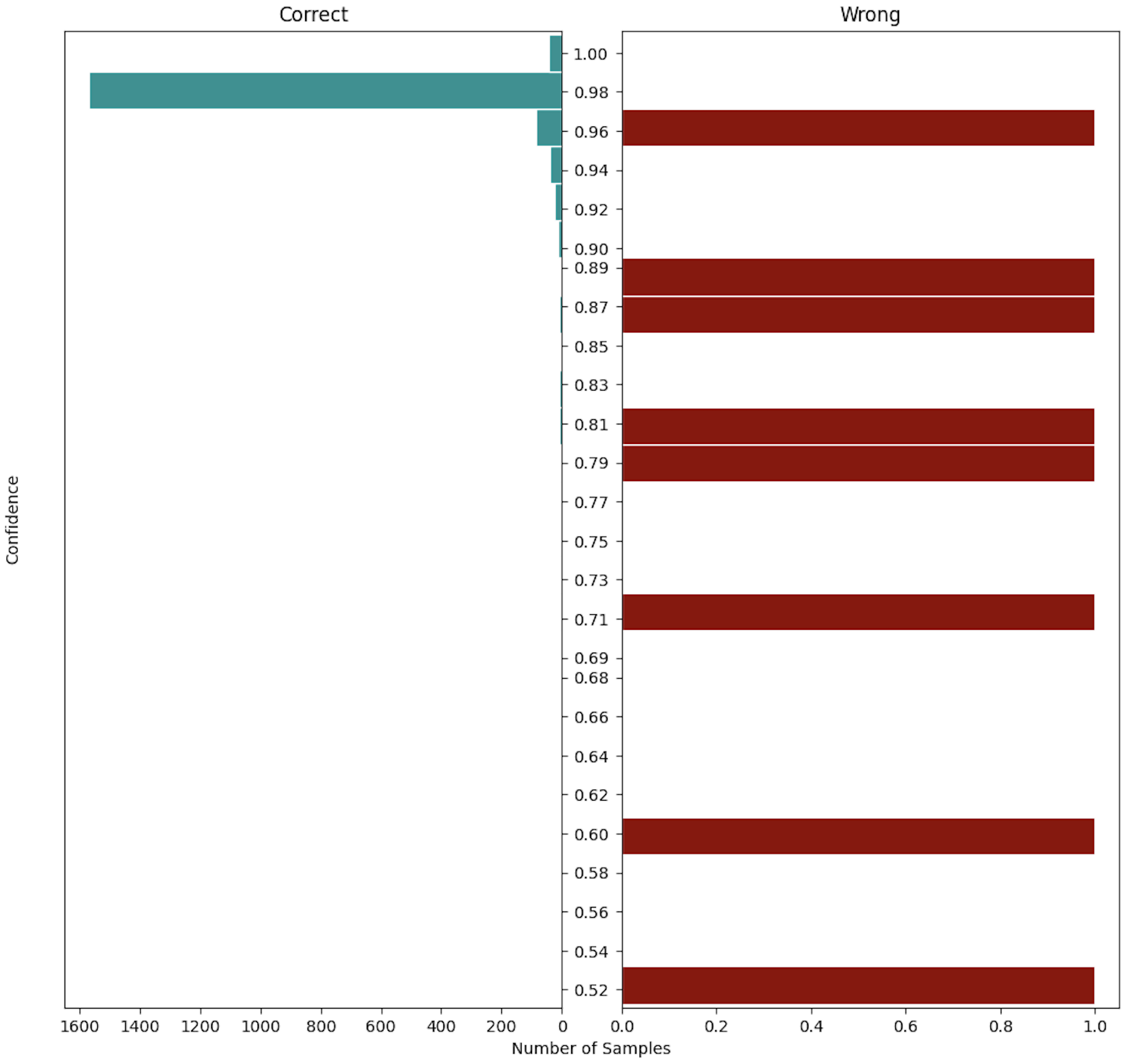}}
\caption{Intent prediction confidence distribution on the Wolof dataset}
\label{wolof_intent}
\end{figure}

The results of the intent classification are shown in Table \ref{tab:intentsf1}. We show the macro F1 scores on the source and synthetic data and we can observe that we get equivalent scores on both sides. This shows that the model succeeds in discriminating the intents in the synthetic dataset, suggesting a sufficiently qualitative translation. However, the model seems less confident when it comes to Wolof predictions, as shown in Fig.\ref{french_intent} and Fig.\ref{wolof_intent}. 

\begin{figure}
\centerline{\includegraphics[width=0.5\textwidth, keepaspectratio]{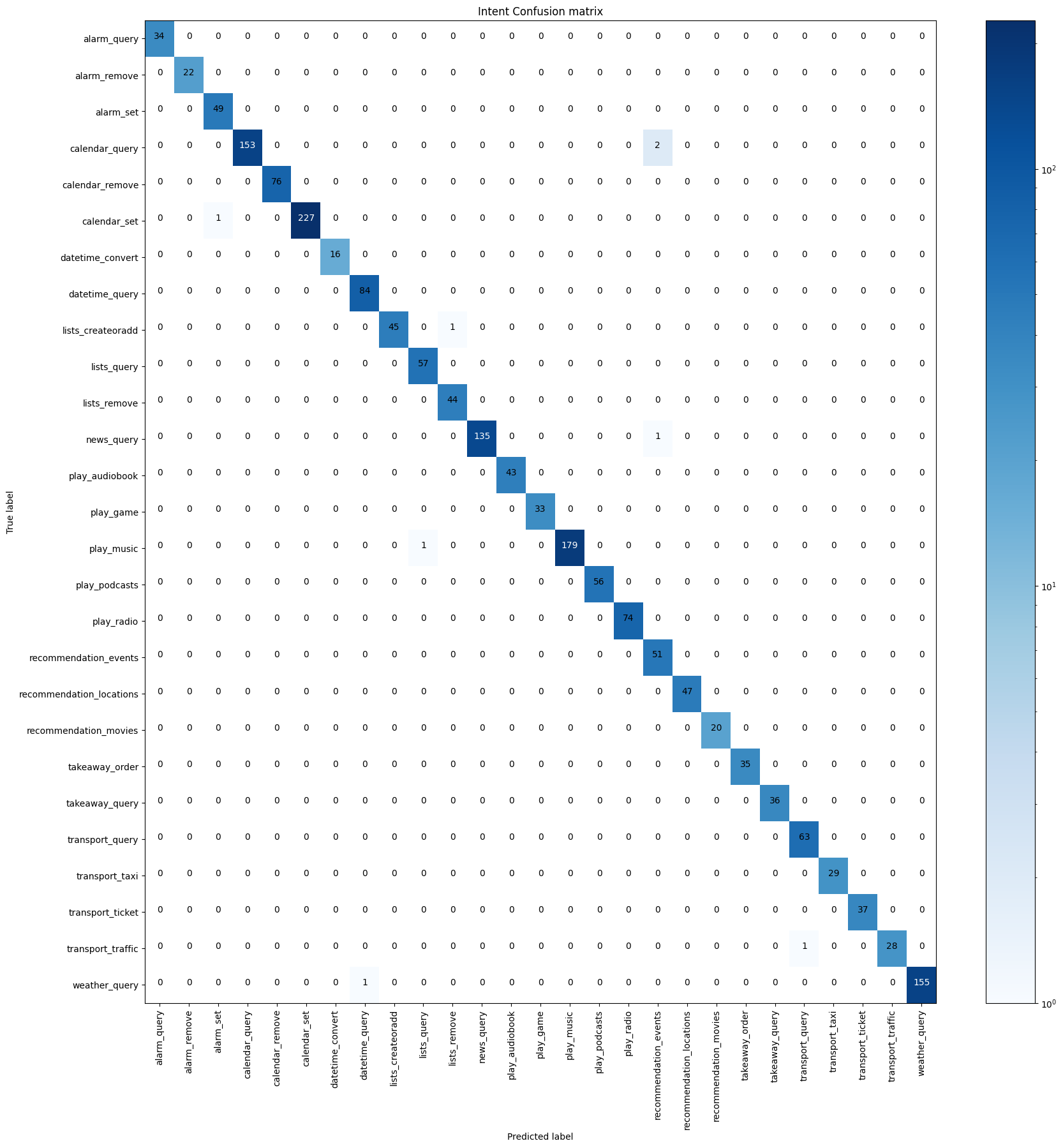}}
\caption{Intent confidence matrix on the Wolof dataset}
\label{wolof_intent_confusion}
\end{figure}

Confidence scores for good predictions vary more in the synthetic dataset than in the source dataset. This shows that, after translation, the model is more likely to confuse certain intentions, as illustrated by Fig.\ref{wolof_intent_confusion}. The \texttt{calendar\_query} and \texttt{reccomandation\_events} intents, for example, show confusion even though they are quite distinct. This can be attributed to the translation system, which may have had difficulty in producing accurate translations in some cases.

\begin{table}[]
\centering
\begin{tabular}{@{}lrl@{}}
\toprule
\multicolumn{1}{c}{\textbf{Intents}} & \multicolumn{1}{c}{\textbf{French}} & \textbf{Wolof} \\ \midrule
\multicolumn{1}{|l|}{transport\_query} & \multicolumn{1}{r|}{1.0} & \multicolumn{1}{l|}{1.0} \\ \midrule
\multicolumn{1}{|l|}{transport\_ticket} & \multicolumn{1}{r|}{1.0} & \multicolumn{1}{l|}{1.0} \\ \midrule
\multicolumn{1}{|l|}{transport\_taxi} & \multicolumn{1}{r|}{1.0} & \multicolumn{1}{l|}{1.0} \\ \midrule
\multicolumn{1}{|l|}{transport\_traffic} & \multicolumn{1}{r|}{\textbf{1.0}} & \multicolumn{1}{l|}{0.98} \\ \midrule
\multicolumn{1}{|l|}{calendar\_query} & \multicolumn{1}{r|}{\textbf{0.99}} & \multicolumn{1}{l|}{\textbf{0.99}} \\ \midrule
\multicolumn{1}{|l|}{calendar\_set} & \multicolumn{1}{r|}{\textbf{1.0}} & \multicolumn{1}{l|}{0.99} \\ \midrule
\multicolumn{1}{|l|}{calendar\_remove} & \multicolumn{1}{r|}{0.99} & \multicolumn{1}{l|}{\textbf{1.0}} \\ \midrule
\multicolumn{1}{|l|}{alarm\_set} & \multicolumn{1}{r|}{\textbf{1.0}} & \multicolumn{1}{l|}{0.98} \\ \midrule
\multicolumn{1}{|l|}{alarm\_remove} & \multicolumn{1}{r|}{1.0} & \multicolumn{1}{l|}{1.0} \\ \midrule
\multicolumn{1}{|l|}{alarm\_query} & \multicolumn{1}{r|}{1.0} & \multicolumn{1}{l|}{1.0} \\ \midrule
\multicolumn{1}{|l|}{lists\_query} & \multicolumn{1}{r|}{\textbf{1.0}} & \multicolumn{1}{l|}{0.99} \\ \midrule
\multicolumn{1}{|l|}{lists\_remove} & \multicolumn{1}{r|}{\textbf{1.0}} & \multicolumn{1}{l|}{0.97} \\ \midrule
\multicolumn{1}{|l|}{lists\_createoradd} & \multicolumn{1}{r|}{\textbf{1.0}} & \multicolumn{1}{l|}{0.98} \\ \midrule
\multicolumn{1}{|l|}{takeaway\_query} & \multicolumn{1}{r|}{1.0} & \multicolumn{1}{l|}{1.0} \\ \midrule
\multicolumn{1}{|l|}{takeaway\_order} & \multicolumn{1}{r|}{1.0} & \multicolumn{1}{l|}{1.0} \\ \midrule
\multicolumn{1}{|l|}{play\_audiobook} & \multicolumn{1}{r|}{1.0} & \multicolumn{1}{l|}{1.0} \\ \midrule
\multicolumn{1}{|l|}{play\_game} & \multicolumn{1}{r|}{1.0} & \multicolumn{1}{l|}{1.0} \\ \midrule
\multicolumn{1}{|l|}{play\_music} & \multicolumn{1}{r|}{1.0} & \multicolumn{1}{l|}{1.0} \\ \midrule
\multicolumn{1}{|l|}{play\_podcasts} & \multicolumn{1}{r|}{1.0} & \multicolumn{1}{l|}{1.0} \\ \midrule
\multicolumn{1}{|l|}{play\_radio} & \multicolumn{1}{r|}{1.0} & \multicolumn{1}{l|}{1.0} \\ \midrule
\multicolumn{1}{|l|}{news\_query} & \multicolumn{1}{r|}{\textbf{1.0}} & \multicolumn{1}{l|}{0.99} \\ \midrule
\multicolumn{1}{|l|}{recommendation\_locations} & \multicolumn{1}{r|}{1.0} & \multicolumn{1}{l|}{1.0} \\ \midrule
\multicolumn{1}{|l|}{recommendation\_events} & \multicolumn{1}{r|}{\textbf{1.0}} & \multicolumn{1}{l|}{0.97} \\ \midrule
\multicolumn{1}{|l|}{recommendation\_movies} & \multicolumn{1}{r|}{1.0} & \multicolumn{1}{l|}{1.0} \\ \midrule
\multicolumn{1}{|l|}{datetime\_query} & \multicolumn{1}{r|}{\textbf{1.0}} & \multicolumn{1}{l|}{0.99} \\ \midrule
\multicolumn{1}{|l|}{datetime\_convert} & \multicolumn{1}{r|}{1.0} & \multicolumn{1}{l|}{1.0} \\ \midrule
\multicolumn{1}{|l|}{weather\_query} & \multicolumn{1}{r|}{1.0} & \multicolumn{1}{l|}{1.0} \\ \midrule
\textbf{macro avg} & \textbf{0.999} & 0.995 \\ \bottomrule
\end{tabular}
\caption{Report of the F1 scores of the intent classification on the French extract from the Massive dataset and its translation into Wolof}
\label{tab:intentsf1}
\end{table}

Table \ref{tab:annotations} illustrates the performance of annotation projection on the two datasets French and Wolof expressed in micro and macro F1 score as well as in accuracy. We note a pronounced discrepancy between the two datasets, with predictions on the synthetic data lower than those on the source data. It is important to point out that some of the annotations in the French dataset are expressions (not words) and are translated separately from the base sentence and therefore from the original context. This can affect the quality of the translation and the ability of the final model to discriminate between different labels. Generally speaking, annotation projection is heavily influenced by the machine translation system, which needs to be optimized as much as possible to create high-quality synthetic datasets.

\begin{table}[]
\centering
\begin{tabular}{@{}lll@{}}
\toprule
 & \textbf{French} & \textbf{Wolof} \\ \midrule
\multicolumn{1}{|l|}{\textbf{micro avg}} & \multicolumn{1}{l|}{0.97} & \multicolumn{1}{l|}{0.89} \\ \midrule
\multicolumn{1}{|l|}{\textbf{macro avg}} & \multicolumn{1}{l|}{0.96} & \multicolumn{1}{l|}{0.86} \\ \midrule
\textbf{accuracy} & 0.98 & 0.94 \\ \bottomrule
\end{tabular}
\caption{Slot filling performance on French and Wolof datasets in micro and macro F1 Score and accuracy}
\label{tab:annotations}
\end{table}

\section{Conclusion and Perspectives}\label{concl}
In this paper, we illustrated an efficient approach to building dialogue systems in a low-resource language. We have shown how to leverage machine translation systems to create synthetic datasets using our annotation projection method. Our experiments showed that training on these synthetic datasets in Wolof gave competitive results compared with the French source data, which is a resource-rich language. However, this approach is strongly affected by the quality of the translation system, and translating annotations out of their original contexts can reduce the final quality of the dataset.
In the future, we will study techniques for improving the quality of the projection, as well as improving the representativeness of Wolof in a language model such as LaBSE. We will also study data augmentation approaches, which are particularly important when addressing low-resource domains. We will also study the integration of spelling correction systems to take into account the variations in writing commonly observed in a language like Wolof.

\section*{Acknowledgment}
This work is supported by the Google PhD Fellowship program.

\bibliography{mybibliography}

\end{document}